\documentclass[twoside]{article}

\usepackage[preprint]{aistats2026}

\usepackage[round]{natbib}

\renewcommand{\cite}{\citep}

\usepackage{amsmath}
\usepackage{amssymb}
\usepackage{amsfonts}
\usepackage{bm}
\usepackage{enumitem}
\usepackage{graphicx}
\usepackage{algorithm}
\usepackage{algorithmic}
\usepackage{subcaption}
\usepackage{multirow}
\usepackage{booktabs}
\usepackage{url}
\newcommand{\Input}{\item[\textbf{Input:}]}
\newcommand{\Output}{\item[\textbf{Output:}]}

\begin{document}

\twocolumn[

\aistatstitle{Hypergraph Neural Networks Accelerate MUS Enumeration}

\aistatsauthor{ Hiroya Ijima \And Koichiro Yawata }

\aistatsaddress{ Hitachi, Ltd. } ]

\begin{abstract}
  Enumerating Minimal Unsatisfiable Subsets (MUSes) is a fundamental task in constraint satisfaction problems (CSPs).
  Its major challenge is the exponential growth of the search space, which becomes particularly severe when satisfiability checks are expensive.
  Recent machine learning approaches reduce this cost for Boolean satisfiability problems but rely on explicit variable-constraint relationships, limiting their application domains.
  This paper proposes a domain-agnostic method to accelerate MUS enumeration using Hypergraph Neural Networks (HGNNs).
  The proposed method incrementally builds a hypergraph with constraints as vertices and MUSes enumerated until the current step as hyperedges, and employs an HGNN-based agent trained via reinforcement learning to minimize the number of satisfiability checks required to obtain an MUS.
  Experimental results demonstrate the effectiveness of our approach in accelerating MUS enumeration, showing that our method can enumerate more MUSes within the same satisfiability check budget compared to conventional methods.
\end{abstract}

\begin{figure*}[t]
  \centering
  \includegraphics[width=\linewidth]{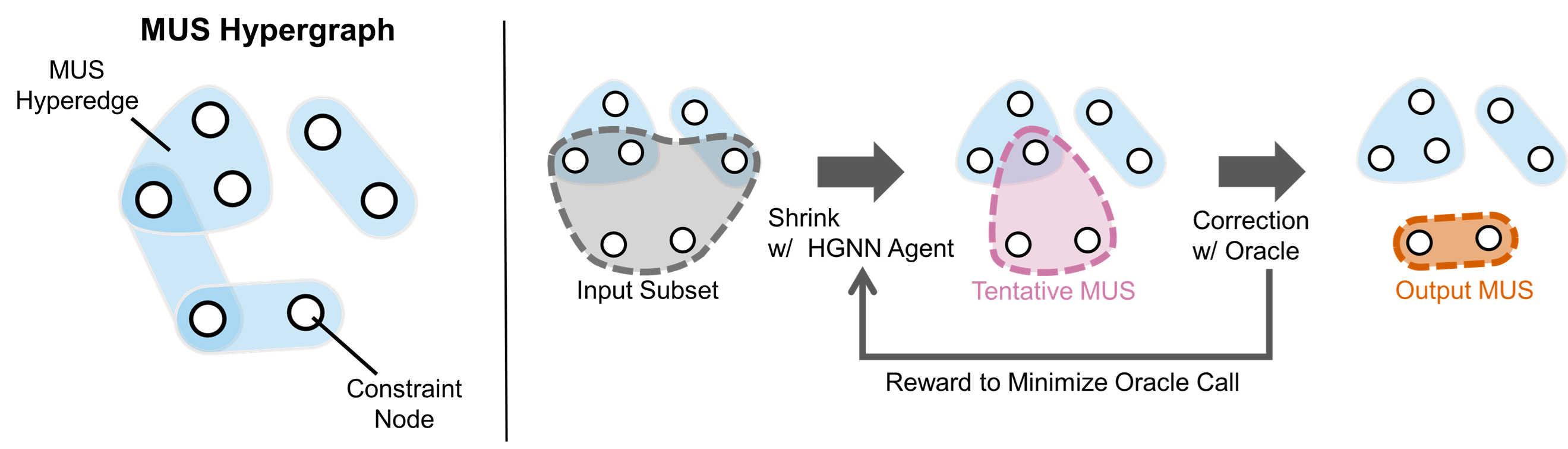}
  \caption{Overview of HyMUSE.
  HyMUSE is a domain-agnostic method to accelerate MUS/MSS enumeration using Hypergraph Neural Networks (HGNNs).
  HyMUSE constructs a hypergraph whose vertices are constraints and whose hyperedges are MUSes and MCSes (only MUSes are illustrated in this figure). The HGNN-based agent takes the hypergraph of MUSes/MCSes enumerated until the current step and the subset of constraints to shrink/grow as input, and outputs a tentative MUS/MSS candidate. The candidate is then corrected to a valid MUS/MSS. The agent is trained via reinforcement learning with reward designed to minimize the number of calls to the satisfiability checking oracle required for the correction.}
  \label{fig:overview}
\end{figure*}

\section{INTRODUCTION}
Minimal Unsatisfiable Subset (MUS) enumeration is a fundamental task of constraint satisfaction problems (CSPs).
An MUS is a subset of constraints that cannot be satisfied simultaneously, and removing any constraint from the subset renders it satisfiable.
Identifying MUSes is crucial for various applications, such as conflict identification in product configurations \cite{herud2022conflict}, inconsistency detection in software requirements \cite{bendik2017consistency} and extracting irreducible infeasible constraint sets (IISes) in optimization problems \cite{chinneck1991locating}.
MUSes are expected to play an increasingly important role with the recent advancements in AI technologies, as they can contribute to explainability for machine learning \cite{ignatiev2019abduction, ignatiev2020contrastive}.
Similarly, Maximal Satisfiable Subsets (MSSes) are important.
An MSS is a subset that can be satisfied simultaneously, and adding any constraint to the subset renders it unsatisfiable.
Since MSSes correspond to Pareto-optimal combinations of constraints, their identification is equally important.

A major challenge of MUS/MSS enumeration is the high computational cost.
The search space of MUS enumeration grows exponentially with respect to the number of constraints, requiring numerous satisfiability checks.
This issue becomes particularly critical when the satisfiability checks are expensive, such as in cases involving complex simulations or large-scale optimization problems.
Efficiently enumerating MUSes/MSSes while minimizing the number of checks is therefore essential for practical applications.

For efficient MUS/MSS enumeration, various algorithms have been proposed.
The main approach for MUS/MSS extraction is the deletion-based approach, which iteratively shrinks an unsatisfiable subset by removing constraints until an MUS is obtained, or grows a satisfiable subset of constraints by adding constraints until an MSS is obtained \cite{chinneck1991locating, bakker1993diagnosing}.
Since enumerating all MUSes is computationally infeasible in many practical cases, online enumeration methods, which output MUSes/MSSes as they are found, are often employed.
Various online enumeration algorithms have been proposed to improve the efficiency of MUS/MSS enumeration by effectively mapping the explored search space and selecting unexplored subsets to shrink/grow \cite{previti2013partial, liffiton2013enumerating, liffiton2016fast, bendik2016tunable, bendik2018recursive}.

Recently, machine learning models have been applied to accelerate MUS enumeration.
For example, NeuroMUSX \cite{moriyama2023gnn} represents variable-constraint relationships in Boolean satisfiability problems as graphs and utilizes graph neural networks (GNNs) to reduce the number of satisfiability checks in deletion operations of MUS extraction by trimming unsatisfiable cores.
GRAPE-MUST \cite{lymperopoulos2024graph} also employs GNNs to narrow down the search space of MUS enumeration.
These methods have shown promising results in accelerating MUS enumeration for Boolean satisfiability problems.

However, existing machine learning-based methods have limitations in their applicability.
They rely on explicit variable-constraint relationships, which are not available in many practical CSPs involving real-valued variables or higher-order constraints.
For such CSPs, constructing variable-constraint graphs is difficult, limiting the applications of existing machine learning-based methods to other domains.

In this paper, we propose a novel domain-agnostic method $\textbf{HyMUSE}$, which accelerates MUS/MSS enumeration using Hypergraph Neural Networks (HGNNs).
Instead of a variable–constraint relation graph, our method constructs a hypergraph representing the constraint–MUS/MSS relations obtained during the enumeration process, with constraints as nodes and incrementally identified MUSes/MCSes as hyperedges.
Given the MUS/MCS hypergraph as input, an HGNN-based agent iteratively deletes/adds constraints to produce a tentative MUS/MSS candidate, which is then corrected to a valid MUS/MSS.
The agent is trained via reinforcement learning to minimize the number of satisfiability checks required for the correction.
Since the MUS/MCS hypergraph is available in any CSP during MUS/MSS enumeration, our method is domain-agnostic.

Experimental results demonstrate the effectiveness of our approach in accelerating MUS/MSS enumeration.
By integrating our method with existing enumeration algorithms, more MUSes/MSSes can be enumerated within the same satisfiability check budget.
The proposed method also exhibits generalization capabilities across different problem distributions from the training data, showing its potential for practical applications.

The main contributions of this paper are as follows:
\begin{itemize}[leftmargin=*, itemsep=5pt, topsep=0pt, parsep=0pt]
  \item We propose a domain-agnostic method for MUS/MSS enumeration that uses hypergraph neural networks (HGNNs) to exploit constraint-MUS/MSS relationships, without relying on the explicit variable-constraint relationships required by existing machine learning approaches.
  \item Our method employs an HGNN-based agent that selects constraints to delete/add during shrink/grow operations, producing a tentative MUS/MSS candidate that is then corrected into a valid solution with few satisfiability checks.
  \item The agent is trained via reinforcement learning to minimize the number of satisfiability checks required for the correction.
  \item Experiments show that our method finds more MUSes/MSSes within the same satisfiability-check budget than conventional methods and generalizes across different problem types.
\end{itemize}

\section{RELATED WORK}
\subsection{MUS Enumeration}
Minimal Unsatisfiable Subset (MUS) enumeration is a task of identifying MUSes in a given set of constraints.
MUS correspond to irreducible explanations of unsatisfiability, and thus MUS enumeration plays a crucial role in various applications \cite{herud2022conflict, bendik2017consistency, ignatiev2019abduction, ignatiev2020contrastive, chinneck1991locating}.
Since the candidate subsets of MUS are $2^{C}$, where $C$ is the set of constraints, the search space grows exponentially with the number of constraints, making MUS enumeration computationally expensive.
This challenge becomes particularly severe when satisfiability checks are expensive, therefore, minimizing the number of checks is desirable.

To enumerate MUSes efficiently, various algorithms have been proposed.
A widely used approach for single MUS/MSS extraction is the deletion-based approach, which iteratively shrinks an unsatisfiable set of constraints by removing constraints until an MUS is obtained, or grows a satisfiable set of constraints by adding constraints until an MSS is obtained \cite{chinneck1991locating, bakker1993diagnosing}.
In many practical scenarios, enumerating all MUSes is computationally infeasible, so online enumeration methods are commonly employed, where MUSes/MSSes are output one by one as they are found.
A representative algorithm in this category is MARCO \cite{previti2013partial, liffiton2013enumerating, liffiton2016fast}, which grows satisfiable sets to MSSes and shrinks unsatisfiable sets to MUSes iteratively while mapping the explored search space using blocking conjunctive normal form (CNF) clauses.
Other notable algorithms include TOME \cite{bendik2016tunable} and ReMUS \cite{bendik2018recursive}, which further improve the efficiency of MUS enumeration by refining the heuristics for selecting subsets to shrink.

Our proposed method can be integrated with these existing algorithms by replacing part of the shrink/grow operations with actions performed by machine learning models.

\subsection{Machine Learning for MUS Enumeration}
Recent advances in machine learning have led to the development of methods that use machine learning models to improve CSP solving, including search heuristics, variable assignment, and satisfiability prediction \cite{popescu2022overview, guo2023machine}.
Many of these methods employ graph neural networks (GNNs) to capture the structure of variable-constraint relationships in CSPs \cite{bunz2017graph, selsam2018learning, selsam2019guiding, wang2021neuroback, li2022nsnet, tonshoff2022one}.
For MUS enumeration, GRAPE-MUST \cite{lymperopoulos2024graph} represents literal-clause relationships in CNFs as graphs and utilizes GNNs to narrow down the search space of MUS enumeration.
NeuroMUSX \cite{moriyama2023gnn} also employs GNNs to reduce the number of satisfiability checks in deletion operations of MUS extraction by trimming unsatisfiable cores.
These methods successfully accelerate MUS enumeration for Boolean satisfiability problems.
However, for many types of CSPs, it is difficult to construct variable-constraint graphs, making it challenging to apply existing machine learning-based methods to other domains.

In contrast, our proposed method is domain-agnostic and can be applied to any CSPs, as it does not rely on explicit variable-constraint relationships.

\subsection{Hypergraph Neural Networks}
Hypergraph Neural Networks (HGNNs) are a class of neural networks designed to operate on hypergraphs.
Hypergraphs are generalizations of graphs where edges, called hyperedges, are allowed to connect any number of vertices, that is, a hyperedge is a subset of vertices.
Many types of real-world data can be naturally represented as hypergraphs, such as social networks, biological networks, and recommendation systems, so HGNNs have been widely studied and applied in various domains \cite{kim2024survey, yang2025recent}.

There have been several types of HGNNs proposed, including methods of expanding hypergraphs to corresponding ordinary graphs and applying GNNs \cite{yadati2019hypergcn, huang2021unignn}, spectral convolutional methods that generalize graph convolution to hypergraphs \cite{feng2019hypergraph, bai2021hypergraph}, message-passing methods that directly define message-passing operations on hypergraphs \cite{dong2020hnhn, arya2020hypersage}, and methods that utilize multi-head attention \cite{zhang2019hyper, bai2021hypergraph, arya2020hypersage, chien2021you}.
Our proposed method employs multi-head attention-based HGNNs following AllSetTransformer \cite{chien2021you}.

\begin{figure*}[t]
  \centering
  \includegraphics[width=\linewidth]{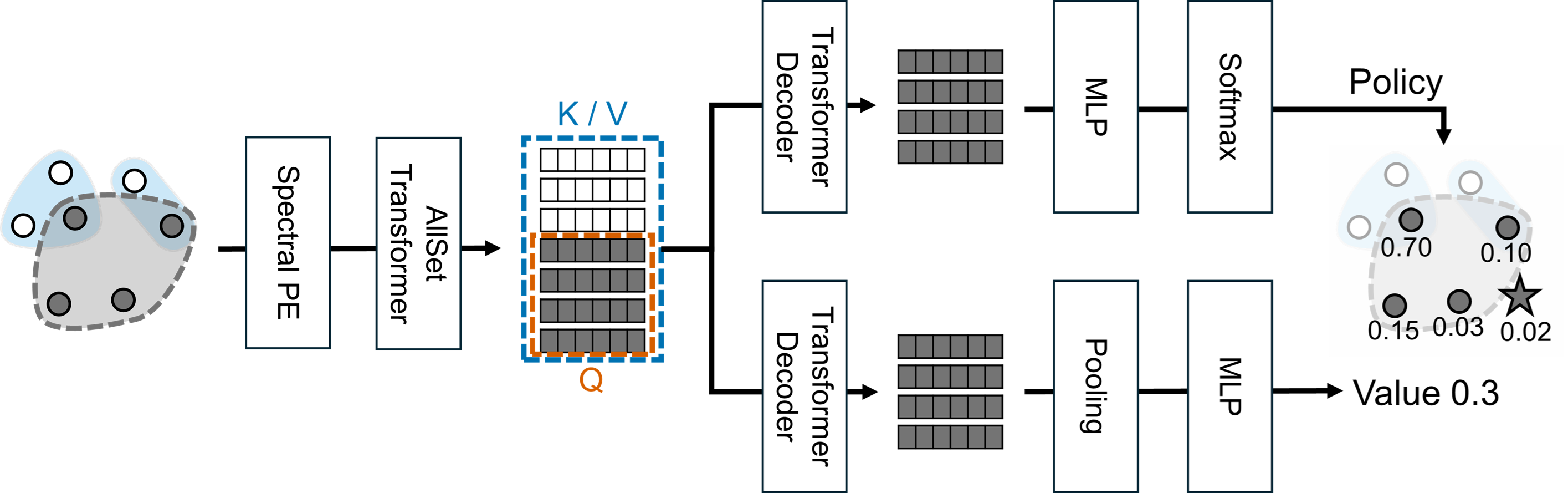}
  \caption{Architecture of Hypergraph Neural Networks. The input is the MUS/MCS hypergraph and the subset of constraints to be shrunk/grown, shown as gray-colored nodes. The feature vectors of vertices are initialized with spectral embeddings based on the hypergraph Laplacian and updated by AllSetTransformer layers. Then, the feature vectors of constraints in the action candidate set are updated separately by two transformer decoders to output the policy and value.}
  \label{fig:hgnn}
\end{figure*}

\subsection{Reinforcement Learning for CSP}
Reinforcement learning (RL) is a framework for sequential decision-making in which an agent learns a policy to maximize rewards through interactions with an environment.
This paradigm has recently been applied to learn search heuristics for solving CSPs \cite{yolcu2019learning, kurin2020can, tonshoff2022one, li2024reinforcement, zhai2025learning}.

In the context of MUS/MSS enumeration, we formulate the selection of constraints to delete/add in shrink/grow operations as a Markov Decision Process (MDP) and train an HGNN-based agent to improve enumeration efficiency.
The agent is trained using Proximal Policy Optimization (PPO) \cite{schulman2017proximal}, a widely used RL algorithm, which has shown strong performance in various applications.

\section{PRELIMINARIES}
We first formalize the problem setting and key concepts used throughout this paper.

Consider a general constraint satisfaction problem (CSP) defined over a finite set of constraints $C = \{c_1, c_2, \ldots, c_m\}$.
For any subset $S \subseteq C$, its satisfiability can be checked by an oracle solver $Oracle: 2^C \rightarrow \{satisfiable, unsatisfiable\}$.

The definitions of MUS, MSS, and MCS are as follows.

\begin{itemize}[leftmargin=*, itemsep=5pt, topsep=0pt, parsep=0pt]
  \item \textbf{Minimal Unsatisfiable Subset (MUS):}\\
  A subset $M \subseteq C$ such that $Oracle(M) = unsatisfiable$ and $\forall c \in M, Oracle(M \setminus \{c\}) = satisfiable$.
  \item \textbf{Maximal Satisfiable Subset (MSS):}\\
  A subset $M \subseteq C$ such that $Oracle(M) = satisfiable$ and $\forall c \in C \setminus M, Oracle(M \cup \{c\}) = unsatisfiable$.
  \item \textbf{Minimal Correction Subset (MCS):}\\
  A subset $M \subseteq C$ such that $Oracle(C \setminus M) = satisfiable$ and $\forall c \in M, Oracle(C \setminus (M \setminus \{c\})) = unsatisfiable$.
  Note that MCS is the complement of MSS and is the hitting set of all MUSes.
\end{itemize}

Deletion-based MUS/MSS extraction algorithms typically consist of two operations:
\begin{itemize}[leftmargin=*, itemsep=5pt, topsep=0pt, parsep=0pt]
  \item \textbf{Shrink:} Given an unsatisfiable subset $S \subseteq C$, iteratively delete constraints from $S$ while maintaining unsatisfiability and output an MUS.
  \item \textbf{Grow:} Given a satisfiable subset $S \subseteq C$, iteratively add constraints to $S$ while maintaining satisfiability and output an MSS.
\end{itemize}
In online MUS/MSS enumeration, a seed subset is picked from the unexplored search space, and then if it is unsatisfiable, the shrink operation is applied to obtain an MUS; if it is satisfiable, the grow operation is applied to obtain an MSS.

\begin{figure*}[t]
\centering
\vspace{-5mm}
\begin{minipage}[t]{0.48\linewidth}
  \begin{algorithm}[H]
  \captionsetup{type=algorithm}
  \captionof{algorithm}{Correction for Shrink}
  \label{alg:error_correction_for_shrink}
  \begin{algorithmic}[1]
  \Input Output subset $S' \subseteq C$ and ordered list of constraints selected to delete $D$ of shrink operation with HGNN agent
  \Output An MUS $M \subseteq C$
  \IF{$Oracle(S') = satisfiable$}
      \STATE $M \leftarrow S'$
      \WHILE{$Oracle(M) = satisfiable$}
          \STATE $c \leftarrow D.pop()$
          \STATE $M \leftarrow M \cup \{c\}$
      \ENDWHILE
      \STATE $M \leftarrow \text{Shrink}(M)$
  \ELSE
      \STATE $M \leftarrow \text{Shrink}(S')$
  \ENDIF
  \RETURN $M$
  \end{algorithmic}
  \end{algorithm}
\end{minipage}
\hfill
\begin{minipage}[t]{0.48\linewidth}
  \begin{algorithm}[H]
  \captionsetup{type=algorithm}
  \captionof{algorithm}{Correction for Grow}
  \label{alg:error_correction_for_grow}
  \begin{algorithmic}[1]
  \Input Output subset $S' \subseteq C$ and ordered list of constraints selected to add $A$ of grow operation with HGNN agent
  \Output An MSS $M \subseteq C$
  \IF{$Oracle(S') = unsatisfiable$}
      \STATE $M \leftarrow S'$
      \WHILE{$Oracle(M) = unsatisfiable$}
          \STATE $c \leftarrow A.pop()$
          \STATE $M \leftarrow M \setminus \{c\}$
      \ENDWHILE
      \STATE $M \leftarrow \text{Grow}(M)$
  \ELSE
      \STATE $M \leftarrow \text{Grow}(S')$
  \ENDIF
  \RETURN $M$
  \end{algorithmic}
  \end{algorithm}
\end{minipage}
\end{figure*}

\section{PROPOSED METHOD}
This work aims to accelerate MUS/MSS enumeration by reducing the number of satisfiability checks through a domain-agnostic machine learning-based approach.
To achieve this, we propose HyMUSE, a method that represents the MUSes/MCSes enumerated up to the current step as hypergraphs and employs an HGNN-based agent trained via reinforcement learning to minimize the number of checks required to correct the outputs into valid MUSes/MSSes.
An overview of the proposed method is shown in Figure~\ref{fig:overview}.

In the proposed method, part of the conventional shrink/grow operations is replaced by actions performed by an HGNN-based agent.
At each enumeration step $t$, when an unexplored subset $S \subseteq C$ is picked, the agent performs the shrink operation to obtain a tentative MUS if $S$ is unsatisfiable, or the grow operation to obtain a tentative MSS if $S$ is satisfiable.
Given the MUS/MCS hypergraph $H^{(t)}$ and the subset $S$ to be shrunk/grown, the agent iteratively selects a constraint to delete/add or decides to finish the operation and outputs the resulting subset.
Since the output subset is not guaranteed to be a valid MUS/MSS, a correction algorithm is applied to convert the tentative subset into valid MUS/MSS with a small number of satisfiability checks.
The HGNN-based agent is trained via reinforcement learning to minimize the number of checks required for the correction procedure.

In the following sections, we describe the details of each component of HyMUSE.
First, we define the MUS/MCS hypergraph.
Then, we explain the architecture of the hypergraph neural network used in the agent and how it outputs the policy for selecting constraints to delete/add.
Next, we describe the correction algorithms.
Finally, we explain the training method of the HGNN-based agent via reinforcement learning.

\begin{figure*}[t]
  \centering
  \vspace{-5mm}
  \begin{minipage}[t]{0.48\linewidth}
    \begin{algorithm}[H]
    \captionsetup{type=algorithm}
    \captionof{algorithm}{Shrink with Agent}
    \label{alg:shrink_with_agent}
    \begin{algorithmic}[1]
    \Input An unsatisfiable set of constraints $S \subseteq C$ and MUS/MCS hypergraph $H^{(t)}$
    \Output An MUS $M \subseteq S$
    \STATE $S_0^{(t)} \leftarrow S$
    \STATE $D \leftarrow [\ ]$
    \FOR{$\tau = 0, 1, 2, \ldots$}
        \STATE $a_{\tau}^{(t)} \sim \pi(a | H^{(t)}, S_{\tau}^{(t)})$ from HGNN model
        \IF{$a_{\tau}^{(t)}$ is $finish$}
            \STATE break
        \ELSE
            \STATE $S_{\tau+1}^{(t)} \leftarrow S_{\tau}^{(t)} \setminus \{a_{\tau}^{(t)}\}$
            \STATE $D.push(a_{\tau}^{(t)})$
        \ENDIF
    \ENDFOR
    \STATE $M \leftarrow \text{CorrectionForShrink}(S_{\tau}^{(t)}, D)$
    \RETURN $M$
    \end{algorithmic}
    \end{algorithm}
  \end{minipage}
  \hfill
  \begin{minipage}[t]{0.48\linewidth}
    \begin{algorithm}[H]
    \captionsetup{type=algorithm}
    \captionof{algorithm}{Grow with Agent}
    \label{alg:grow_with_agent}
    \begin{algorithmic}[1]
    \Input A satisfiable set of constraints $S \subseteq C$ and MUS/MCS hypergraph $H^{(t)}$
    \Output An MSS $M \supseteq S$
    \STATE $S_0^{(t)} \leftarrow S$
    \STATE $A \leftarrow [\ ]$
    \FOR{$\tau = 0, 1, 2, \ldots$}
        \STATE $a_{\tau}^{(t)} \sim \pi(a | H^{(t)}, S_{\tau}^{(t)})$ from HGNN model
        \IF{$a_{\tau}^{(t)}$ is $finish$}
            \STATE break
        \ELSE
            \STATE $S_{\tau+1}^{(t)} \leftarrow S_{\tau}^{(t)} \cup \{a_{\tau}^{(t)}\}$
            \STATE $A.push(a_{\tau}^{(t)})$
        \ENDIF
    \ENDFOR
    \STATE $M \leftarrow \text{CorrectionForGrow}(S_{\tau}^{(t)}, A)$
    \RETURN $M$
    \end{algorithmic}
    \end{algorithm}
  \end{minipage}
\end{figure*}

\subsection{MUS/MCS Hypergraph}
Our method constructs hypergraphs where vertices represent constraints and hyperedges represent obtained MUSes/MCSes until the current step.
In online MUS/MSS enumeration, we can obtain an MUS or an MSS at each enumeration step. 
In our method, let $MUS^{(t)} = \{M_1, M_2, \ldots, M_k\}$ be the set of MUSes enumerated until step $t$, and $MCS^{(t)} = \{M'_1, M'_2, \ldots, M'_l\}$ be the set of MCSes obtained as complements of MSSes enumerated until step $t$.
Considering each constraint $c \in C$ as a vertex and each MUS $M_i \in MUS^{(t)}$ and MCS $M'_j \in MCS^{(t)}$ as hyperedges, the exploration state at step $t$ can be represented as a heterogeneous hypergraph $H^{(t)} = (\mathcal{V}, \mathcal{E}^{(t)}_{MUS}, \mathcal{E}^{(t)}_{MCS})$ where $\mathcal{V} = C$ is the set of vertices representing constraints, and $\mathcal{E}^{(t)}_{MUS} = MUS^{(t)}$ and $\mathcal{E}^{(t)}_{MCS} = MCS^{(t)}$ are the sets of hyperedges representing MUSes and MCSes, respectively.
Since the MUS/MCS hypergraph is available in any CSP during MUS/MSS enumeration, even when variable-constraint relationships are unknown, it is a domain-agnostic representation of CSPs.

\subsection{Hypergraph Neural Networks}
The HGNN takes the MUS/MCS hypergraph $H^{(t)}$ and the subset $S \subseteq C$ of constraints to shrink/grow as input, and outputs the policy $\pi(a | H^{(t)}, S)$ for action $a$ that selects a constraint to delete/add and the value $V(H^{(t)}, S)$ for training with reinforcement learning.
Our model employs AllSetTransformer-style set-to-set attention \cite{chien2021you} to embed the structure of MUS/MCS hypergraph.
The architecture of our model is shown in Figure~\ref{fig:hgnn}.

In the model, the feature vector of each vertex $v \in \mathcal{V}$ is initialized with spectral embedding derived from the normalized hypergraph Laplacian proposed by \citet{zhou2006learning}.
Then, the vertex features are updated by AllSetTransformer layers.
In each layer, the feature vectors are updated separately on MUS hyperedges and MCS hyperedges using different parameters and then combined by linear projection.
\begin{eqnarray}
h_v^{(l+1)} = W^{(l)} \cdot \left( \left(f^{(l)}_{\mathcal{E}^{(t)}_{MUS} \rightarrow \mathcal{V}} \circ f^{(l)}_{\mathcal{V} \rightarrow \mathcal{E}^{(t)}_{MUS}}\right)(h_v^{(l)}) \nonumber \right. \\
\left. \| \left(f^{(l)}_{\mathcal{E}^{(t)}_{MCS} \rightarrow \mathcal{V}} \circ f^{(l)}_{\mathcal{V} \rightarrow \mathcal{E}^{(t)}_{MCS}}\right)(h_v^{(l)}) \right)
\end{eqnarray}
Here, $h_v^{(l)} \in \mathbb{R}^d$ is the feature vector of vertex $v$ in the $l$-th layer, $W^{(l)} \in \mathbb{R}^{2d \times d}$ is a learnable weight matrix, $\circ$ is the function composition operation, and $\|$ is the concatenation operation.
$f^{(l)}_{\mathcal{V}\rightarrow \mathcal{E}}$ and $f^{(l)}_{\mathcal{E} \rightarrow \mathcal{V}}$ are AllSetTransformer operations defined by \citet{chien2021you}, which are multi-head attention-based set-to-set functions that aggregate feature vectors of vertices $\mathcal{V}$ to hyperedges $\mathcal{E}$ and vice versa.

After the AllSetTransformer layers, the candidate set is extracted and fed into two separate transformer decoders \cite{vaswani2017attention} to output the policy and value.
When performing shrink, the candidate set is the input subset $S$, and when performing grow, the candidate set is the complement of the input subset $C \setminus S$.
In both transformer decoders, the embeddings of vertices in the candidate set are used as queries, while all vertex embeddings are used as keys/values in cross-attention, enabling the model to consider the similarities of substructures in the MUS/MCS hypergraph.

To compute the policy $\pi(a | H^{(t)}, S)$, the output feature vectors of the candidate constraints from the policy decoder are passed through several feed-forward layers, producing a logit for each constraint in the candidate set. A virtual $finish$ action with a fixed logit of zero is appended, and softmax is applied over all logits to obtain the action probability distribution.
To compute the value $V(H^{(t)}, S)$, the output feature vectors from the value decoder are aggregated via average and max pooling, concatenated, and passed through several feed-forward layers to produce a scalar estimate.

Because HGNN-based agent acts based on domain-agnostic MUS/MCS hypergraph, our method can be applied regardless of the types of variables and constraints, as long as the satisfiability can be checked by an oracle solver.

\subsection{Correction Algorithm}
The output subset of shrink/grow operations with the HGNN agent is not guaranteed to be a valid MUS/MSS due to prediction errors.
The correction algorithm converts the tentative subsets into valid MUSes/MSSes with satisfiability check oracle.
The algorithms for shrink/grow operations are shown in Algorithm~\ref{alg:error_correction_for_shrink} and Algorithm~\ref{alg:error_correction_for_grow}.

For shrink correction, if the output subset $S$ is satisfiable, the deleted constraints are restored one by one in reverse order of their deletion until $S$ becomes unsatisfiable again, and then the standard shrink algorithm is applied to obtain an MUS.
If the output subset $S$ is already unsatisfiable, the standard shrink algorithm is directly applied to ensure minimality.

For grow correction, if the output subset $S$ is unsatisfiable, the added constraints are removed one by one in reverse order of their addition until $S$ becomes satisfiable again, and then the standard grow algorithm is applied to obtain an MSS.
If the output subset $S$ is already satisfiable, the standard grow algorithm is directly applied to ensure maximality.

\subsection{Training}
The HGNN agent is trained via reinforcement learning to minimize the number of satisfiability checks required for the correction procedure.
We formulate the iterative selection of constraints to delete/add in shrink/grow operations as a Markov Decision Process (MDP) and train the model using Proximal Policy Optimization (PPO) \cite{schulman2017proximal}.
The shrink and grow operations with the agent are shown in Algorithm~\ref{alg:shrink_with_agent} and Algorithm~\ref{alg:grow_with_agent}.

The state $s_{\tau}^{(t)}$ at step $\tau$ in the shrink/grow operation at enumeration step $t$ is defined as the pair of the MUS/MCS hypergraph $H^{(t)}$ and the subset $S_{\tau}^{(t)} \subseteq C$.
For the shrink operation, $S_{0}^{(t)}$ is initialized to an unexplored unsatisfiable subset of constraints. At each step $\tau$, the agent chooses an action $a_{\tau}^{(t)} \in S_{\tau}^{(t)} \cup \{finish\}$: either selecting a constraint $c \in S_{\tau}^{(t)}$ to remove from the subset, or terminating the operation by choosing $finish$.
For the grow operation, $S_{0}^{(t)}$ is initialized to an unexplored satisfiable subset of constraints. At each step $\tau$, the agent chooses an action $a_{\tau}^{(t)} \in (C \setminus S_{\tau}^{(t)}) \cup \{finish\}$: either selecting a constraint $c \in C \setminus S_{\tau}^{(t)}$ to add to the subset, or terminating the operation by choosing $finish$.

The transition is deterministic, and the next state $s_{\tau+1}^{(t)}$ is obtained by deleting/adding the selected constraint $c$ from/to $S_{\tau}^{(t)}$.
The episode ends when the action $finish$ is selected, and the reward $r^{(t)}$ is given at the end of the episode.
The reward $r^{(t)}$ for the shrink/grow operation is calculated as follows:
\begin{eqnarray}
r^{(t)} = \begin{cases}
\displaystyle 1 - \frac{N_{correction} - |MUS|}{|S_0^{(t)}|} & \text{shrink} \\
\\
\displaystyle 1 - \frac{N_{correction} - |C \setminus MSS|}{|C \setminus S_0^{(t)}|} & \text{grow}
\end{cases}
\end{eqnarray}
Here, $N_{correction}$ is the number of satisfiability checks required for the correction procedure after the shrink/grow operation.
$|MUS|$ and $|MSS|$ are the sizes of the valid MUS and MSS obtained after the correction, respectively.
The minimum of $N_{correction}$ is $|MUS|$ when shrinking and $|C \setminus MSS|$ when growing, which is achieved when the output of the agent is already a valid MUS/MSS and no correction is needed.
The reward is designed to encourage the agent to minimize the number of checks required for the correction.

We train the model by collecting episodes of shrink/grow operations with the current policy and updating the model parameters to maximize the expected reward via PPO.

\section{EXPERIMENTS}
We evaluated HyMUSE to address the following research questions:
\begin{enumerate}[label=\textbf{RQ\arabic*:}, leftmargin=0.7cm, align=left, nosep]
  \item Does HyMUSE accelerate MUS/MSS enumeration compared to conventional methods?
  \item Does HyMUSE generalize to different problem distributions from those seen during training?
  \item How does HyMUSE perform when integrated with different enumeration algorithms?
\end{enumerate}

\subsection{Training Settings}
\begin{figure}[t]
  \centering
  \includegraphics[width=0.8\linewidth]{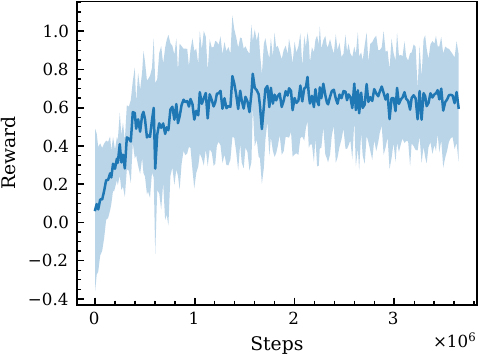}
  \caption{Reward Transition during Training. The average reward and the standard deviation are plotted.}
  \label{fig:reward_plot}
\end{figure}

The agent was trained on CNF instances drawn from $\mathbf{SR}(\mathbf{U}(5, 20))$, which is the distribution of random CNF instances introduced by \citet{selsam2018learning}.
In this distribution, the number of variables is uniformly sampled from 5 to 20, and each constraint is then generated by randomly sampling a subset of variables and negating each variable with probability 0.5.
The training steps were performed in combination with MARCO \cite{previti2013partial, liffiton2013enumerating, liffiton2016fast} to select unexplored sets to shrink/grow.
In total, training comprised 3.6M steps.
Figure~\ref{fig:reward_plot} shows the reward curve during training, indicating that the reward increased as training progressed and eventually converged.

\subsection{Evaluation Settings}
We evaluated the performance of the proposed method with the trained agent on four datasets of different problem distributions.
\begin{itemize}[leftmargin=*, itemsep=5pt, topsep=0pt, parsep=0pt]
  \item $\bm{SAT_{small}}$: 500 CNF instances drawn from  $\mathbf{SR}(\mathbf{U}(5, 20))$, the same distribution as in training.
  \item $\bm{SAT_{large}}$: 500 CNF instances drawn from $\mathbf{SR}(\mathbf{U}(20, 40))$ with larger numbers of variables and constraints than in $SAT_{small}$.
  \item $\bm{GC}$: 500 CNF instances derived from graph coloring problems randomly generated with the method described in \citet{lymperopoulos2024graph}.
  \item $\bm{SMT}$: 495 Satisfiability Modulo Theories (SMT) instances from the benchmark in SMT-LIB \cite{BarFT-SMTLIB, preiner2025smtlib}, referring to \citet{bendik2018recursive}.
  Note that these SMT instances include problems that cannot be converted to CNF, making it impossible to represent variable-constraint relationships as graphs and apply existing GNN-based methods.
\end{itemize}

For each dataset, we integrated the trained agent with three different enumeration strategies MARCO \cite{previti2013partial, liffiton2013enumerating, liffiton2016fast}, TOME \cite{bendik2016tunable} and ReMUS \cite{bendik2018recursive}, and compared performance with and without the agent.
The main evaluation metric is the total number of MUSes and MSSes enumerated within a fixed number of satisfiability checks.
This metric is appropriate for evaluating the efficiency of enumeration, as the cost of satisfiability checks dominates the overall computational cost of enumeration and is independent of the implementation and hardware.
Only those instances that completed the fixed number of checks within 600 seconds and without error were included in the evaluation.

\subsection{Results}
The results corresponding to each research question are presented below.
\subsubsection{Acceleration of MUS Enumeration}\label{sec:acceleration_results}
\begin{figure}[t]
  \centering
  \includegraphics[width=0.65\linewidth,]{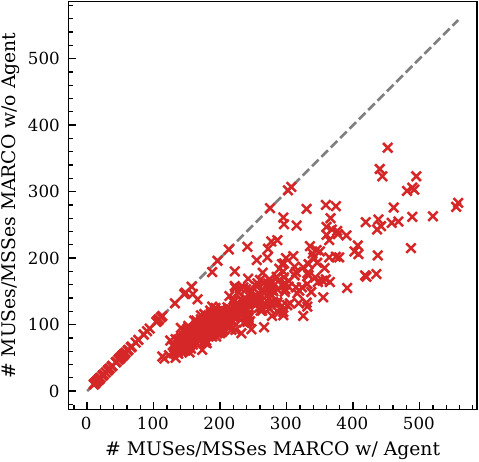}
  \caption{Comparison of MARCO with and without the Agent on $SAT_{small}$. The x-axis shows the number of MUSes/MSSes found with the agent, and the y-axis shows the number found without the agent within 10,000 checks. Points below the dashed $y=x$ line indicate improvement by the agent. Points on the line are instances where both methods found all MUSes/MSSes.}
  \label{fig:marco_agentic_vs_non_agentic_sr5-20}
\end{figure}

Figure~\ref{fig:marco_agentic_vs_non_agentic_sr5-20} compares MARCO with and without the agent on $SAT_{small}$ under a limit of 10,000 satisfiability checks.
As shown in the figure, MARCO with the agent enumerated more MUSes/MSSes than MARCO without the agent on all instances except those for which both methods completed the enumeration of all MUSes/MSSes.

\begin{figure}[t]
  \centering
  \includegraphics[width=0.8\linewidth]{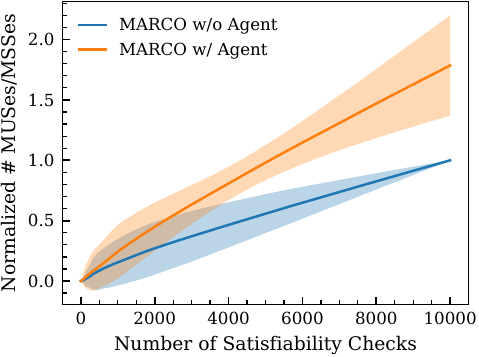}
  \caption{Relation of Number of MUSes/MSSes and Number of Satisfiability Checks. The number of MUSes/MSSes is plotted against the number of satisfiability checks for both MARCO with and without the agent.}
  \label{fig:mus_mss_num_transition}
\end{figure}

Figure~\ref{fig:mus_mss_num_transition} shows the cumulative number of enumerated MUSes/MSSes as a function of the number of satisfiability checks.
The numbers of MUSes/MSSes are normalized by that of MARCO without the agent at 10,000 checks for each instance.
Table~\ref{tab:acceleration_results} shows the improvement ratio in the number of MUSes/MSSes when integrating the trained agent with MARCO on $SAT_{small}$.
The improvement ratio is calculated as the number of MUSes/MSSes enumerated with the agent divided by that without the agent at each instance and averaged across instances.
The improvement ratio consistently exceeds 1 across different numbers of checks and quartiles.

\begin{table}[t]
\caption{Improvement Ratio of MARCO with the Agent on $SAT_{small}$. The improvement ratio is calculated as the number of MUSes/MSSes with the agent divided by that without the agent at each instance and averaged across instances in each quartile group and overall. The quartiles are based on the number of MUSes/MSSes enumerated without the agent.}
\label{tab:acceleration_results}
\begin{center}
\begin{tabular}{cccc}
\toprule
 Quartile & 1k checks & 5k checks & 10k checks \\
\midrule
1Q & 1.80 $\pm$ 0.57 & 2.05 $\pm$ 0.55 & 1.73 $\pm$ 0.58 \\
2Q & 1.78 $\pm$ 0.47 & 1.90 $\pm$ 0.40 & 1.89 $\pm$ 0.33 \\
3Q & 1.77 $\pm$ 0.36 & 1.82 $\pm$ 0.31 & 1.86 $\pm$ 0.33 \\
4Q & 1.57 $\pm$ 0.29 & 1.61 $\pm$ 0.31 & 1.67 $\pm$ 0.33 \\
\midrule
Overall & 1.74 $\pm$ 0.45 & 1.85 $\pm$ 0.43 & 1.79 $\pm$ 0.42 \\
\bottomrule
\end{tabular}
\end{center}
\end{table}

\subsubsection{Performance on Different Distributions}
Table~\ref{tab:generalization_results} shows the improvement ratio when integrating the trained agent with MARCO on datasets with different distributions from the training distribution.
Integrating the agent improved the number of MUSes/MSSes enumerated on all datasets.
A large variation in performance is observed on $GC$ and $SMT$.

\begin{table}[t]
\caption{The Number of MUSes/MSSes with and without the Agent and the Improvement Ratio on Different Datasets at 10k Checks. The number of MUSes/MSSes and the improvement ratio are averaged across instances in each dataset and shown with standard deviation.}
\label{tab:generalization_results}
\begin{center}
\setlength{\tabcolsep}{3pt}
\begin{tabular}{cccc}
\toprule
Dataset & \multicolumn{2}{c}{\# MUSes/MSSes} & Ratio \\
\midrule
\multirow{2}{*}{$SAT_{small}$} & w/o Agent & 119.5 $\pm$ 60.3 & \multirow{2}{*}{1.79 $\pm$ 0.42} \\
& w/ Agent & $\bm{211.3}$ $\pm$ $\bm{99.4}$ & \\
\midrule
\multirow{2}{*}{$SAT_{large}$} & w/o Agent & 54.1 $\pm$ 15.3 & \multirow{2}{*}{2.46 $\pm$ 0.42} \\
& w/ Agent & $\bm{130.6}$ $\pm$ $\bm{31.6}$ & \\
\midrule
\multirow{2}{*}{$GC$} & w/o Agent & 15.1 $\pm$ 16.1 & \multirow{2}{*}{2.32 $\pm$ 0.96} \\
& w/ Agent & $\bm{31.6}$ $\pm$ $\bm{28.4}$ & \\
\midrule
\multirow{2}{*}{$SMT$} & w/o Agent & 43.3 $\pm$ 100.7 & \multirow{2}{*}{1.30 $\pm$ 0.62} \\
& w/ Agent & $\bm{58.0}$ $\pm$ $\bm{103.2}$ & \\
\bottomrule
\end{tabular}
\end{center}
\end{table}

\subsubsection{Performance with Different Enumeration Algorithms}
Table~\ref{tab:combination_results} shows the improvement ratio in the number of MUSes/MSSes when integrating the trained agent with different enumeration algorithms from the algorithms used during training.
As shown in the table, the improvement is marginal when integrating the agent with TOME and ReMUS, in contrast to the significant improvement observed with MARCO.

\section{DISCUSSION}
The results demonstrate that our method successfully accelerates MUS enumeration, producing significant improvements in the number of MUSes/MSSes enumerated within a fixed number of satisfiability checks.

The results on different distributions indicate that our method generalizes well to different problem distributions, which is a significant advantage for practical use. 
The results on $SMT$ indicate that our method is domain-agnostic and can be applied to various types of CSPs, including those where explicit variable-constraint relationships are not available.
The large variation in performance on different distributions from the training distribution suggests that the agent learned both domain-agnostic heuristics effective across different distributions and some heuristics specific to particular problem distributions.

The results on different enumeration algorithms suggest that the agent trained with MARCO specializes for optimizing MARCO operations and does not generalize well to other methods.
This is presumably because the TOME and ReMUS have different strategies for efficient selection of unexplored sets to shrink/grow compared to MARCO and thus the agent has little margin to improve the performance.
Training the agent in combination with each enumeration algorithm would yield better performance.
The result of training with ReMUS is provided in Appendix~\ref{appendix:training_with_different_enumeration_algorithms}.

It would be interesting as future work to induce the acquisition of more general heuristics by training the agent on more diverse problem distributions and with different enumeration algorithms.
Though MUS and MCS are used as hyperedges in this work, it is important to compare the performance among different combinations of subsets used as hyperedges, such as using only MUS, only MCS, or using other subsets of constraints including MSS and complements of MUS.
It is also expected that observing the behavior of the trained agent and analyzing the HGNN using XAI methods will lead to the discovery of new rule-based heuristics.

\section{CONCLUSION}
In this paper, to accelerate MUS enumeration in domain-agnostic CSPs, we proposed a hypergraph neural network-based method HyMUSE.
Our method constructs hypergraphs where vertices represent constraints and hyperedges represent MUSes/MCSes enumerated until the current step and employs HGNN-based agents to obtain tentative MUS/MSS candidates.
The HGNN-based agents are trained via reinforcement learning to minimize the number of satisfiability checks required to correct the candidates to valid MUSes/MSSes.
Experimental results demonstrate the effectiveness of our method in accelerating MUS enumeration, producing significant improvements in the number of MUSes/MSSes enumerated within a fixed number of satisfiability checks.
The results also show that our method generalizes well to different problem distributions, indicating its potential for practical applications.

\begin{table}[t]
\caption{Improvement Ratio by Integrating the Agent with Different Enumeration Methods on $SAT_{small}$. The improvement ratio is calculated as the number of MUSes/MSSes found by each enumeration method with the agent divided by that without the agent at each instance and averaged across instances.}
\label{tab:combination_results}
\begin{center}
\begin{tabular}{lccc}
\toprule
Method & 1k checks & 5k checks & 10k checks \\
\midrule
TOME & 0.92 $\pm$ 0.14 & 1.08 $\pm$ 0.18 & 1.11 $\pm$ 0.16 \\
ReMUS & 1.31 $\pm$ 0.29 & 1.05 $\pm$ 0.37 & 0.75 $\pm$ 0.29 \\
\bottomrule
\end{tabular}
\end{center}
\end{table}

\section*{Acknowledgements}
The authors are grateful to Kota Dohi and Taiki Morinaga for their constructive comments and suggestions on the paper,
and to Yoshihiro Osakabe and Akinori Asahara for their guidance and continuous support for the research.

\bibliography{aistats2026.bib}

\clearpage
\appendix
\section*{Appendix}
\section{Theoretical Performance Analysis}
\subsection{Best and Worst Case Performance}
The standard shrink and grow procedures described in \citet{liffiton2016fast} are shown in Algorithm~\ref{alg:shrink} and Algorithm~\ref{alg:grow}.
The number of satisfiability checks required for the standard shrink/grow algorithms is $|S|$ and $|C \setminus S|$, respectively, where $S$ is the input subset of constraints to be shrink/grow.

\begin{algorithm}[h]
\caption{Shrink}
\label{alg:shrink}
\begin{algorithmic}[1]
\Input An unsatisfiable subset of constraints $S \subseteq C$
\Output An MUS $M \subseteq S$
\STATE $M \leftarrow S$
\FOR{each constraint $c$ in $S$}
    \IF{$Oracle(M \setminus \{c\}) = satisfiable$}
        \STATE continue
    \ELSE
        \STATE $M \leftarrow M \setminus \{c\}$
    \ENDIF
\ENDFOR
\RETURN $M$
\end{algorithmic}
\end{algorithm}

\begin{algorithm}[h]
\caption{Grow}
\label{alg:grow}
\begin{algorithmic}[1]
\Input A satisfiable subset of constraints $S \subseteq C$
\Output An MSS $M \supseteq S$
\STATE $M \leftarrow S$
\FOR{each constraint $c$ in $C \setminus S$}
    \IF{$Oracle(M \cup \{c\}) = unsatisfiable$}
        \STATE continue
    \ELSE
        \STATE $M \leftarrow M \cup \{c\}$
    \ENDIF
\ENDFOR
\RETURN $M$
\end{algorithmic}
\end{algorithm}

In our method, the minimum number of satisfiability checks required to obtain an MUS/MSS from an unsatisfiable/satisfiable set of constraints $S$ is $|MUS|$ and $|C \setminus MSS|$, respectively, which is achieved when the output of the agent is already a valid MUS/MSS and no correction is needed.

In the worst case, the number of satisfiability checks required for the correction procedure is $2|S|$ when shrinking, which occurs when the agent deletes all constraints in $S$ and the correction procedure needs to add back all constraints.
When growing, the worst case is $2|C \setminus S|$, which occurs when the agent adds all constraints in $C \setminus S$ and the correction procedure needs to remove all added constraints.

\subsection{Computational Cost}
To assess the practical impact of the improvement, comparing the computational cost or time required to enumerate a certain number of MUSes/MSSes with and without the agent is also important.
The effectiveness of the improvement in terms of computational cost depends on the balance between the cost of satisfiability checks and the cost of the agent's inference.
Let $C_{check}$ be the average cost of a satisfiability check, $C_{infer}$ be the average cost of the agent's inference, $N_{check}$ be the number of satisfiability checks required for extraction of an MUS/MSS without the agent, $N'_{check}$ be the number of satisfiability checks required for extraction of an MUS/MSS with the agent, and $N_{infer}$ be the number of times the agent is invoked during extraction.
Our method is effective in terms of computational cost when the following inequality holds:
\begin{eqnarray}
C_{check} \cdot N'_{check} + C_{infer} \cdot N_{infer} < C_{check} \cdot N_{check}
\end{eqnarray}
The left-hand side of the inequality represents the total cost of extraction with the agent, which consists of the cost of satisfiability checks required for correction and the cost of agent inference.
The right-hand side represents the cost of extraction without the agent, which consists of the cost of satisfiability checks required for the standard shrink/grow procedure.
This can be rewritten as:
\begin{eqnarray}
\frac{C_{infer}}{C_{check}} < \frac{N_{check} - N'_{check}}{N_{infer}}
\end{eqnarray}
As shown in the equation, our method is effective when the ratio of the inference cost to the check cost is less than $r_{eff} = (N_{check} - N'_{check})/N_{infer}$, which represents the average number of checks saved per agent invocation.
Table~\ref{tab:effective_ratio} shows the average number of checks required to extract an MUS/MSS with and without the agent, the average number of inference calls, and the effective ratio $r_{eff}$ on each dataset.

\begin{table}[ht]
\caption{Average Number of Checks and Inference Calls for Extracting an MUS/MSS with and without the Agent and the Effective Ratio $r_{eff}$ on Different Datasets. The numbers are averaged across instances in each dataset and shown with standard deviation.}
\label{tab:effective_ratio}
\begin{center}
\resizebox{\columnwidth}{!}{%
\begin{tabular}{ccccc}
\toprule
Dataset & $N_{check}$ & $N'_{check}$ & $N_{infer}$ & $r_{eff}$ \\
\midrule
$SAT_{small}$ & 85 $\pm$ 37 & 44 $\pm$ 17 & 48 $\pm$ 24 & 0.83 $\pm$ 0.11 \\
$SAT_{large}$ & 196 $\pm$ 51 & 81 $\pm$ 22 & 127 $\pm$ 37 & 0.90 $\pm$ 0.08 \\
$GC$ & 996 $\pm$ 540 & 478 $\pm$ 312 & 673 $\pm$ 422 & 0.77 $\pm$ 0.27 \\
$SMT$ & 136 $\pm$ 313 & 117 $\pm$ 428 & 94 $\pm$ 243 & 0.25 $\pm$ 0.71 \\
\bottomrule
\end{tabular}%
}
\end{center}
\end{table}

\section{Details of Experimental Settings}
\subsection{Training Settings}
The agent was trained on randomly generated CNF instances drawn from $\textbf{SR}(\textbf{U}(5, 20))$ distribution, which was introduced by \citet{selsam2018learning}.
The geometric distribution's probability parameter was set to 0.3 as used in \citet{lymperopoulos2024graph}.
At each parameter update, approximately 18k steps were collected through shrink/grow episodes across four different CNF instances using the current policy within 5000 satisfiability checks budget, and the model parameters were updated for four epochs.
In total, training comprised 3.6M steps.
The training steps were performed in combination with MARCO \cite{previti2013partial, liffiton2013enumerating, liffiton2016fast} for selecting unexplored sets to shrink/grow.
The HGNN architecture used three AllSetTransformer layers and three transformer decoder layers, with feature dimension 64 and four attention heads.
Optimization was performed using Adam \cite{kingma2014adam} with a learning rate of $2 \times 10^{-5}$ and a batch size of 1024 via gradient accumulation.

\subsection{Implementation Details}
For implementation of HGNN model, we used PyTorch \cite{paszke2019pytorch} and PyTorch Geometric \cite{fey2019fast}.
For the oracle solver for CNF instances and the solver for mapping clauses in enumeration algorithms, we used PySAT \cite{imms-sat18, itk-sat24} with CaDiCaL \cite{BiereFallerFazekasFleuryFroleyks-CAV24} and Glucose \cite{audemard2009glucose} backend.
For the oracle solver for SMT instances, we used PySMT \cite{pysmt2015} with Z3 \cite{de2008z3} backend.
All experiments were conducted on a machine with an Intel Xeon Gold 6130 CPU with 256GB RAM and four NVIDIA Tesla V100 GPUs with 32GB memory.
Code for reproducing the result in section~\ref{sec:acceleration_results} is available at \url{https://github.com/hitachi-ais/HGNN-MUSE}.

\section{Effect of Training}
To evaluate the effect of training, we compared the performance of the agent trained with MARCO and that of a randomly initialized agent without training when integrated with MARCO on $SAT_{small}$.
Table~\ref{tab:training_effect_results} shows the improvement ratio in the number of MUSes/MSSes.
The trained agent enumerated significantly more MUSes/MSSes than the untrained agent, demonstrating the effectiveness of training.

\begin{table}[h]
\caption{The Improvement Ratio in the Number of MUSes/MSSes when Integrating Trained and Untrained Agents with MARCO on $SAT_{small}$. The numbers are averaged across instances and shown with standard deviation.}
\label{tab:training_effect_results}
\begin{center}
\setlength{\tabcolsep}{3pt}
\begin{tabular}{cccc}
\toprule
 Agent & 1k checks & 5k checks & 10k checks \\
\midrule
Trained & 1.74 $\pm$ 0.45 & 1.85 $\pm$ 0.43 & 1.79 $\pm$ 0.42 \\
Untrained & 1.02 $\pm$ 0.22 & 0.93 $\pm$ 0.15 & 0.88 $\pm$ 0.11 \\
\bottomrule
\end{tabular}
\end{center}
\end{table}

\section{Direct Comparison with Other Enumeration Methods}
To directly compare the performance of different enumeration methods, we compared the number of MUSes/MSSes enumerated within a fixed number of satisfiability checks with and without the agent trained with MARCO.
Table~\ref{tab:normalized_mus_mss_num} shows the normalized number of MUSes/MSSes with different enumeration methods, where the numbers are normalized by that of MARCO without the agent at each instance and averaged across instances.
While ReMUS without the agent performs the best among all combinations of enumeration methods and agent integration, MARCO with the agent outperforms TOME.
It is a significant result considering that TOME has an efficient strategy for selecting unexplored sets to shrink/grow.

\begin{table}[h]
\caption{Normalized Number of MUSes/MSSes Enumerated with Different Enumeration Methods on $SAT_{small}$. The numbers are normalized by that of MARCO without the agent at each instance and averaged across instances.}
\label{tab:normalized_mus_mss_num}
\begin{center}
\setlength{\tabcolsep}{3pt}
\begin{tabular}{cccc}
\toprule
Method & MARCO & TOME & ReMUS \\
\midrule
w/o Agent & (1.00) & 1.52 $\pm$ 0.48 & $\mathbf{3.29 \pm 1.54}$ \\
w/ Agent & 1.79 $\pm$ 0.42 & 1.69 $\pm$ 0.57 & 2.16 $\pm$ 0.58 \\
\bottomrule
\end{tabular}
\end{center}
\end{table}

\section{Training with Different Enumeration Algorithms}\label{appendix:training_with_different_enumeration_algorithms}
We also trained the HGNN-based agent with ReMUS.
ReMUS recursively searches for smaller unexplored subsets to shrink minimizing the number of satisfiability checks required to obtain an MUS and achieves state-of-the-art efficiency on some CSP domains among domain-agnostic MUS enumeration algorithms \cite{bendik2018evaluation}.
Table~\ref{tab:remus_combination_results} and Table~\ref{tab:remus_combination_results_2} show the number of MUSes/MSSes enumerated within 10,000 satisfiability checks for each method with and without the agent trained with ReMUS on $SAT_{small}$ and $SAT_{large}$, respectively.
The numbers are normalized by that of MARCO without the agent at each instance.
The results show that the proposed method improves the performance of ReMUS on $SAT_{large}$, achieving the best result among the evaluated method combinations on this dataset.
However, the performance deteriorates on $SAT_{small}$.
It is assumed that ReMUS already employs a sophisticated heuristic to select small unexplored subsets to shrink, and therefore the agent has limited room for improvement when the scale of the problem is small.

\begin{table}[h]
\caption{Normalized Number of MUSes/MSSes Enumerated within 10,000 Satisfiability Checks on $SAT_{small}$ with and without the Agent Trained with ReMUS.}
\label{tab:remus_combination_results}
\begin{center}
\setlength{\tabcolsep}{3pt}
\begin{tabular}{cccc}
\toprule
Method & MARCO & TOME & ReMUS \\
\midrule
w/o Agent & (1.00) & 1.51 $\pm$ 0.48 & $\mathbf{3.26 \pm 1.48}$ \\
w/ Agent & 1.85 $\pm$ 0.44 & 1.67 $\pm$ 0.56 & 2.19 $\pm$ 0.59 \\
\bottomrule
\end{tabular}
\end{center}
\end{table}

\begin{table}[h]
\caption{Normalized Number of MUSes/MSSes Enumerated within 10,000 Satisfiability Checks on $SAT_{large}$ with and without the Agent Trained with ReMUS.}
\label{tab:remus_combination_results_2}
\begin{center}
\setlength{\tabcolsep}{3pt}
\begin{tabular}{cccc}
\toprule
Method & MARCO & TOME & ReMUS \\
\midrule
w/o Agent & (1.00) & 2.17 $\pm$ 0.40 & 2.00 $\pm$ 0.73 \\
w/ Agent & 2.50 $\pm$ 0.43 & 2.47 $\pm$ 0.47 & $\mathbf{3.01 \pm 0.41}$ \\
\bottomrule
\end{tabular}
\end{center}
\end{table}

\section{Performance Comparison on Different Distributions}
\begin{figure}[t]
  \centering
  \includegraphics[width=0.65\linewidth,]{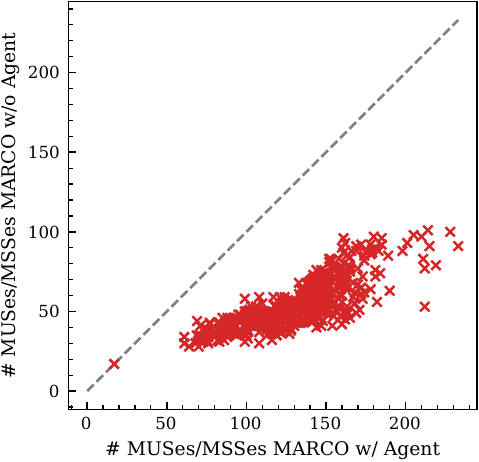}
  \caption{Comparison of MARCO with and without the Agent on $SAT_{large}$.}
  \label{fig:marco_agentic_vs_non_agentic_sr20-40}
\end{figure}

\begin{figure}[t]
  \centering
  \includegraphics[width=0.65\linewidth,]{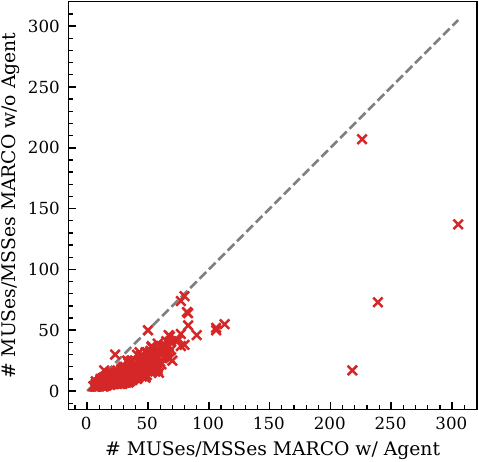}
  \caption{Comparison of MARCO with and without the Agent on $GC$.}
  \label{fig:marco_agentic_vs_non_agentic_gc}
\end{figure}

\begin{figure}[t]
  \centering
  \includegraphics[width=0.65\linewidth,]{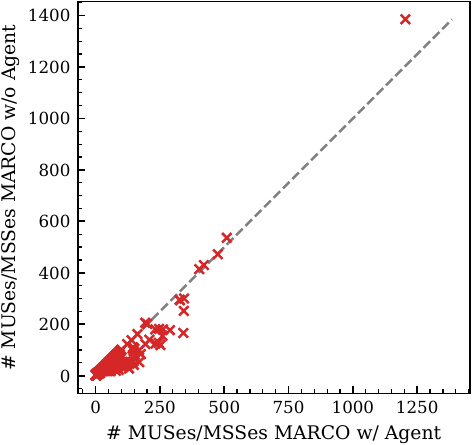}
  \caption{Comparison of MARCO with and without the Agent on $SMT$.}
  \label{fig:marco_agentic_vs_non_agentic_smt}
\end{figure}

The performance comparison of MARCO with and without the agent on $SAT_{large}$, $GC$, and $SMT$ is shown in Figure~\ref{fig:marco_agentic_vs_non_agentic_sr20-40}, Figure~\ref{fig:marco_agentic_vs_non_agentic_gc}, and Figure~\ref{fig:marco_agentic_vs_non_agentic_smt}, respectively.
On $SAT_{large}$, MARCO with the agent enumerated more MUSes/MSSes than MARCO without the agent on almost all instances except for one instance where both methods enumerated all MUSes/MSSes.
Even on $GC$ and $SMT$, where the performance improvement is relatively small compared to $SAT_{small}$, MARCO with the agent outperformed MARCO without the agent on most instances.

\end{document}